\documentclass[10pt,twocolumn,letterpaper]{article}

\usepackage{cvm}
\usepackage{times}
\usepackage{epsfig}
\usepackage{graphicx}
\usepackage{amsmath}
\usepackage{amssymb}

\usepackage{url}
\usepackage{cite}
\usepackage{fancybox}

\usepackage{framed,multirow}
\usepackage{latexsym}
\usepackage{booktabs}
\usepackage{xcolor}
\definecolor{newcolor}{rgb}{.8,.349,.1}

\usepackage{algorithm}
\usepackage{algorithmic}
\usepackage{bm}
\usepackage{pbox}

\usepackage[pagebackref=true,breaklinks=true,letterpaper=true,colorlinks,bookmarks=false]{hyperref}

\cvmfinalcopy
\def\cvmPaperID{****} % *** Enter the cvm Paper ID here

\ifcvmfinal\fi
\begin{document}

%%%%%%%%% TITLE
\title{Deep Tiny Network for Recognition-Oriented Face Image Quality Assessment}

\author{Baoyun Peng\\
Academy of Military Sciences\\
 Beijing, China\\
\and
Min Liu\\
Academy of Military Sciences\\
Beijing, China\\
\and
Zhaoning Zhang\\
National University of Defense Technology\\
Changsha, China\\
\and
Kai Xu\\
National University of Defense Technology\\
Changsha, China\\
\and
Dongsheng Li\\
National University of Defense Technology\\
Changsha, China\\
}

\maketitle

\begin{abstract}
Face recognition has made significant progress in recent years due to deep convolutional neural networks (CNN). In many face recognition (FR) scenarios, face images are acquired from a sequence with huge intra-variations. These intra-variations, which are mainly affected by the low-quality face images, cause instability of recognition performance. Previous works have focused on ad-hoc methods to select frames from a video or use face image quality assessment (FIQA) methods, which consider only a particular or combination of several distortions. 
In this work, we present an efficient non-reference image quality assessment for FR that directly links image quality assessment (IQA) and FR. More specifically, we propose a new measurement to evaluate image quality without any reference. Based on the proposed quality measurement, we propose a deep Tiny Face Quality network (tinyFQnet) to learn a quality prediction function from data. 
We evaluate the proposed method for different powerful FR models on two classical video-based (or template-based) benchmarks: IJB-B and YTF. Extensive experiments show that, although the tinyFQnet is much smaller than the others, the proposed method outperforms state-of-the-art quality assessment methods in terms of effectiveness and efficiency.
\end{abstract}

\section{Introduction}

\begin{figure}[htbp]
  \centering 
  \includegraphics[width=0.45\textwidth]{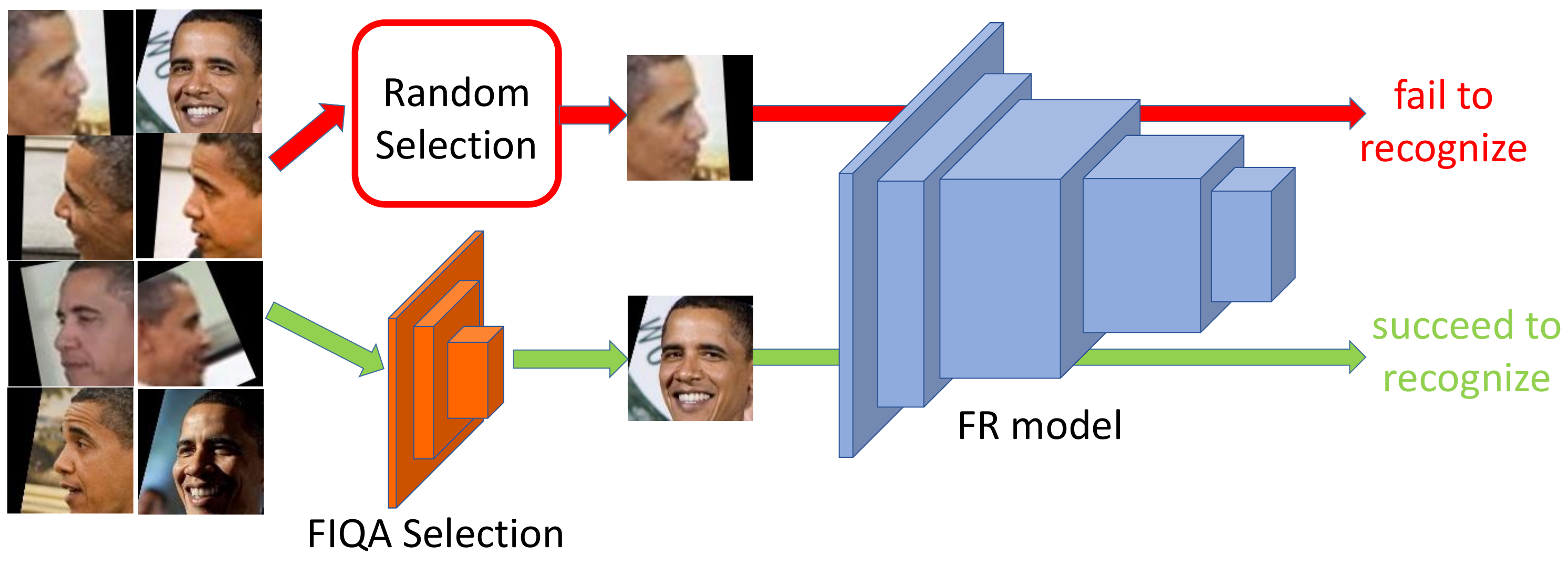}
  \caption{A typical selection process. Random selection may select a low-quality image that is hard for the FR model to recognize. A good FIQA selection method can improve the performance of recognition. }\label{fig:image_selection}
\end{figure}

The performance of face recognition (FR) has been considerably improved in recent years, mainly owing to the combination of deep neural networks and large-scale labeled face images. On the representative academic benchmarks IJB-C \cite{maze2018iarpa} and IQIYI-VID \cite{liu2018iqiyi}, several FR methods \cite{deng2019arcface,huang2020curricularface,an2021partial} have even surpassed humans in terms of face verification. However, in many real scenarios where face images are captured as a sequence with high uncertainty, the FR performance degrades sharply due to the sequence's low-quality face images. Figure \ref{fig:image_selection} shows such a typical scenario in which the input of the FR model is a sequence of images. In such a sequence, the difficulty of recognizing the person in various face images is different. Usually, images with a high-resolution, neural head pose, and little distortions (occlusion, blur, or noises), are easier to recognize. For the sake of throughput capacity, the FR system may only sample one or a few images from a sequence to recognize. Without a good selection method, the FR system may fail to recognize when selecting a low-quality image that is hard to recognize. Face image quality assessment (FIQA) can correctly assess face images' quality to improve an FR system's performance.

Evaluating the influence of face image quality on an FR system's performance is non-trivial since the performance is affected by many underlying variations, and there is no unified definition or standard metric on face image quality. Several efforts have been made to develop common standards \cite{gao2007standardization,sang2009face}. In these standards, the factors that influence the face image quality can be categorized into perceptual variations and biometric variations. 

Many prior works have contributed to automatic FIQA methods \cite{Zhu2011RRAR,abaza2014design,best2018learning}, and these works are similar to general image quality assessment that relies on either the difference among several known properties of the human visual system between target and reference images \cite{eskicioglu1995image} or the degradation in structural information of face images \cite{wang2004image}. However, considering one or several particular factors may result in unsuitable face image selection since the face image quality is influenced by many potential factors.

Several new learning-based FIQA methods have been proposed in recent years that attempt to address limitations of earlier techniques through automatically learning how to assess face image quality from amounts of data \cite{sun2017face,hernandez2019faceqnet,Meng_2021_CVPR,boutros2023cr}. For instance, \cite{vignesh2015face} uses the matching score between a face image and a reference image as a quality score based on hand-crafted features, while \cite{hernandez2019faceqnet} and \cite{Meng_2021_CVPR} are based on features extracted from a well-performed FR model. Both of them need to select an image as a reference for each class to evaluate quality scores before training their models. For non-reference FIQA, \cite{huber2022evaluating} proposes pixel-level face image quality assessment that measures face recognition performance changes when inserting or deleting pixels based on their predicted quality. \cite{chen2021lightqnet} uses knowledge distillation on difficult samples to train an efficient lightweight FIQA network focused on the quality classification boundary. 
Besides, there are also several non-reference IQA methods \cite{babnik2022faceqan,boutros2023cr} that use GAN or CNN to assess face image quality based on noise exploration or sample relative classifiability.

\begin{figure*}[!htb]
	\centering 
	\includegraphics[width=0.95\textwidth]{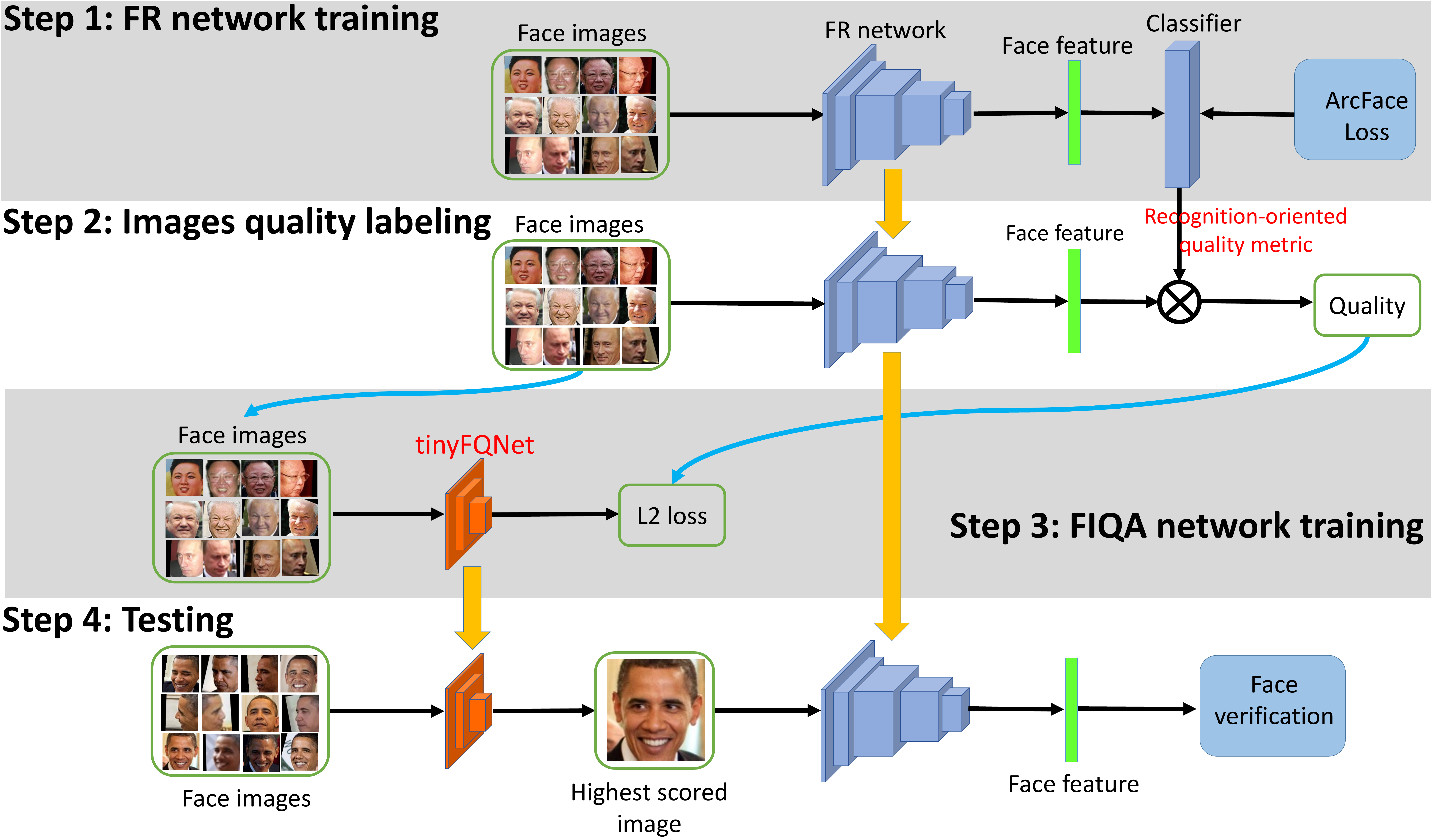}
	\caption{The pipeline for video-based FR. The base set contains multiple identities, and each identity is with several images. In common, the frames are captured under significant variations, such as large head pose, illumination, motion blur, and occlusion. Our method consists of four steps. In Step 1, we train the FR network. Then, the trained FR network is used to label face images with quality. We use L2 regression loss in Step 3 to train a FIQA network. For testing in Step 4, we first use our trained FIQA network to select high-quality face images and feed them to our trained FR network to extract features, then the extracted feature will be used for face verification.}\label{fig:three_comp}
\end{figure*}

% {\color{red} 
Unlike the above FIQA methods, we propose an efficient deep image quality assessment method for face recognition in this paper. More specifically, we propose a novel non-reference quality measurement for face image quality scoring that directly links Face IQA with FR without a reference.
% } 
The details of the proposed quality metric are presented in Section 3.1.
% \ref{sec:quality_metric}.
% {\color{red} 
Using the proposed quality measurement, we can generate amounts of training data with quality labels in a fully automatic way. Besides, previous methods are usually based on complicated networks (e.g., ResNet-50 in \cite{zhuang2019recognition} and VGG-16 in \cite{hernandez2019faceqnet}) while training them on a relatively small dataset. They may ignore that computation and memory costs are important when applying face IQA to a real FR system.
% }
However, the computation or memory cost of FIQA is truly a key factor when designing a resource-limited FR system, such as face unlock on mobile phones and other embedded devices. Based on this consideration, we propose a tiny but effective FIQA network that only has \textbf{21.8k} parameters, and the average time of processing an image is only 4ms on Samsung S10. 

To verify the effectiveness of the proposed method, the performance evaluation of the FR model when combined with face IQA as a plug-in is carried out. We provide extensive analysis of the impact on the performance of the FR model in terms of different network sizes and show that even a tiny network can achieve comparable performance with much more complicated networks on face image quality assessment. To evaluate the influence of different factors on the FR model's performance, we provide extensive experiments on three perceptual factors: head pose, blurriness, and JPEG compression rate. We also provide a comprehensive comparison among perceptual-based and learning-based FIQA methods and compare the proposed method with several other FIQA methods. 

The proposed method can also be used to generate video representation by quality-weighted averages like \cite{yang2016neural,Sankaran2018Metadata}, but this is not our scope in this work. We evaluate the proposed method on three common datasets, including the IARPA Janus Benchmark-B (IJB-B) \cite{whitelam2017iarpa}, IARPA Janus Benchmark-C (IJB-C) \cite{maze2018iarpa}, and Youtube Faces Database (YTF) \cite{Wolf2011Face}. The results show that our proposed method can improve the baseline FR performance significantly and is also superior to other FIQA methods. The contributions of this paper are summarized as follows:

%% introduce our contributions
\begin{itemize}
\item we propose a new quality measurement that directly targets face recognition for face image quality assessment (FIQA), and no reference is required. Using the proposed quality measurement, we can automatically generate amounts of data with quality labels to train the IQA model;
\item Based on our quality measurement, we propose an efficient and effective deep network as a face IQA model and demonstrate the effectiveness of the proposed method on the unconstrained IJB-B dataset and YTF dataset;
\item we provide extensive analysis about the influence of several perceptual factors on FR's performance, and a comprehensive comparison between perceptual-based and learning-based FIQA methods;
\item we provide extensive analysis on the influence of FIQA network size on FR's performance, and show that even a tiny network with only 21.8K parameters can achieve comparable performance with a much more complicated network with 23M parameters;
\item we provide extensive analysis on the influence of data size and quality score distribution of training dataset on FR's performance, and provide an effective sampling method to make the training dataset more balanced, which is proved that it can further improve FR's performance;
\end{itemize}

The rest of this paper is organized as follows. In Section \ref{relate_work}, we review related works on FR and FIQA. In Section \ref{sec:section3}, we present the details of the proposed FIQA method, including a novel recognition-oriented quality metric in section \ref{sec:quality_metric}, a tiny network for FIQA in section \ref{sec:tinyFQnet}, and an easy and efficient  method that can quickly generate amounts of labeled data in section \ref{sec:labelling_data_generation}. In Section \ref{experiment}, evaluations on IJB-B and YTF datasets are presented. Finally, the concluding remarks are presented in Section \ref{conclusion}.

\section{Related Work}\label{relate_work}

This work mainly focuses on face image quality assessment (FIQA) for face recognition. Many of previous works are extended from general image quality assessment (IQA). Therefore, we first discuss the related works of general IQA, then face recognition and FIQA.

\subsection{Image Quality Assessment}
Image quality assessment methods are mainly categorized into full-reference methods and non-reference (or blind) methods. The former needs full access to the reference images \cite{Eckert1998Perceptual,Sheikh2006An,Ke2016Analysis}. Several full-reference IQA methods are based on human visual systems \cite{Eckert1998Perceptual,Eric2010Most}, which predict quality scores from visible image differences.
Changes of structure in a distorted image are also taken to measure the quality in \cite{wang2004image,Sampat2009Complex}.
Non-reference IQA methods require no or limited information about the reference image.
These methods try to detect a particular or several distortions, such as blurriness \cite{Ferzli2009A,Hassen2010no}, JPEG compression \cite{Shan2009No,Zhang2011Kurtosis}, noise \cite{Corner2003noise}, and combination of several distortions \cite{Erez2010No,Moorthy2011Blind}. Recently, several methods \cite{Li2016No,Zhang2020blind,liu2022liqa} have adopted a deep convolutional neural network (CNN) to predict image quality on training datasets with quality labels. Note that the ultimate goal of these quality assessment measurements is for human perceptual cognition rather than face recognition.

\subsection{Face Image Quality Assessment}
Existing methods for face image quality assessment are mainly based on the similarity to a reference (ideal) image, and these methods measure the face image quality by comparing several known properties between target images and reference images. Specifically, perceptual image quality (such as contrast, resolution, sharpness, and noise) and biometric quality (such as pose, illumination, and occlusion) have been used to evaluate face image quality \cite{zhang2009asymmetry,abaza2014design}. Prior works \cite{nasrollahi2008face} proposed a weighted quality fusion approach that merged the weights of factors (rotation, sharpness, brightness, and resolution) into a quality score. Wong et al. \cite{wong2011patch} predicted a face image quality score by determining its probabilistic similarity to an ideal face image via local patch-based analysis for FR in a video. 

Another way for FIQA is to leverage learning-based approaches \cite{best2018learning,chen2015face,vignesh2015face}. 
Unlike conventional approaches that measure the quality by analyzing pre-defined biometric and perceptual image characteristics, learning-based methods learn a prediction function from amounts of face images with quality scores. 
Hence, the learning process is highly dependent on the training dataset. To assess a face image's quality, \cite{Hsu2006Quality, kim2015face} used a discrete value to indicate matching results. Best-Rowden et al. \cite{best2018learning} obtained a training dataset through pairwise comparisons performed by workers from Amazon Mechanical Turk. They used a support vector machine to predict the quality score using features extracted from a face image. Chen et al. \cite{chen2015face} learned a ranking function for quality scores by dividing datasets according to quality.  These methods aim to establish a function from image features to quality scores. Besides, these learning-based methods use either hand-crafted features or features extracted from pre-trained recognition models as inputs to learn the prediction function and train the FIQA model on small datasets collected in laboratories (e.g., FRGC \cite{phillips2000feret}). In \cite{sun2017face}, deep CNN is used to determine the category and degree of degradation in a face image by considering five perceptual image characteristics (resolution, blurriness, additive white Gaussian noise, salt-and-pepper noise, and Poisson noise). In \cite{zhuang2019recognition} and \cite{hernandez2019faceqnet}, the face recognition model is also used to generate quality labels, and the deep CNN model is learned from labeled data to predict quality scores. FaceQAN \cite{babnik2022faceqan} links face image quality with adversarial examples learned by gradient descent information from FR model. Recently, \cite{boutros2023cr} proposes CR-FIQA that estimates the face image quality by predicting its relative classifiability using the sample feature representation with respect to its class center and the nearest negative class center.

\section{Method}\label{sec:section3}

In this section, we describe the details of our proposed FIQA method, including a recognition-oriented non-reference quality metric, a tiny but efficient deep network (tinyFQnet), a simple technique to quickly generate amounts of quality-labeled data that can be used to train the tinyFQnet, and a data sampling method to make the distribution of scores in training dataset more balanced. 

\subsection{Recognition-oriented Non-Reference Quality Measurement} \label{sec:quality_metric}
Empirically, focusing on one particular or a combination of several factors (blurriness, head pose, noise, etc.) may not be effective since FR is affected by amounts of variations, and some of these variations are not quantifiable or hand-crafted. Besides, it is hard to determine the relative importance of each factor. We argue that the QA model should be linked with the face recognition process directly. 

Inspired by \cite{hernandez2019faceqnet}, of which the quality score of a given image is defined as the normalized Euclidean distance between it and a reference image of the same subject, we propose a recognition-oriented non-reference quality measurement that directly links quality assessment with face recognition. More specifically, given an image $\bm{x}$ and its label $y$, the quality score of $\bm{x}$ is defined as the cosine similarity between $\bm{x}$ and the distribution center of the same class with $\bm{x}$ in angular space, as follows:
\begin{equation}
    quality(x) = cosine(\bm{f}_x, \bm{u}_y) = \frac{\bm{f}_{\bm{x}} \cdot \bm{u}_y}{ \left \| \bm{f}_x \right \|   \left \| \bm{u}_y \right \|   } \label{equa:fiqa_metric}
\end{equation}
\noindent where $\bm{f}_{\bm{x}}$ is the feature vector of image $\bm{x}$ extracted by the FR model, $\bm{u}_y$ is the center feature vector of $y$ class.
Compared with using a hand-selected image \cite{zhuang2019recognition} or the highest-quality image selected by a commercial system \cite{hernandez2019faceqnet} as the reference to compute image quality, the proposed method in Equation \ref{equa:fiqa_metric} is more reasonable in the following aspect: $cosine(\bm{f}_x, \bm{u}_y)$ directly reflects how difficult for the FR model to recognize the image $\bm{x}$ correctly. The lower the $cosine(\bm{f}_x, \bm{u}_y)$ is, the more difficult the FR model to recognize $\bm{x}$ correctly.

Although the center $\bm{u}_{y}$ is hard to obtain due to the ground-truth distribution of a given class being unknown for a particular FR model, we can regard the last fully connected layer output of an FR model as the class center in the angular space as in \cite{Liu2016ICML,Liu2017SphereFace}.

\begin{figure}[!htb]
	\centering 
	\includegraphics[width=0.95\linewidth]{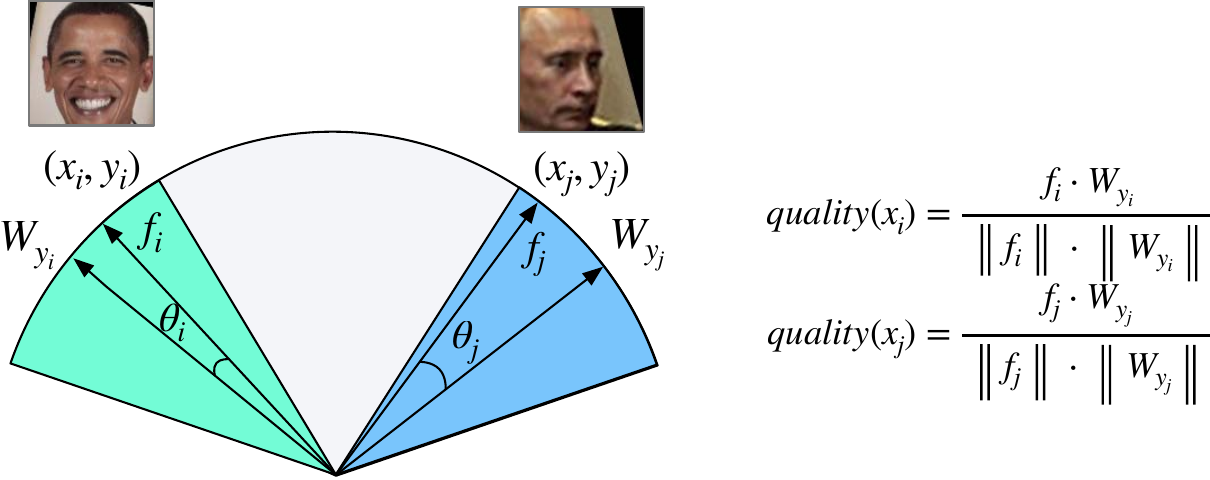}
	\caption{Geometry interpretation of recognition-oriented quality metric in a 2D feature embedding space. $\bm{W}_{y_i}$ and $\bm{W}_{y_j}$ are the centers of $y_i$ and $y_j$ class, $\bm{f}_i$ and $\bm{f}_j$ is the feature vector of $\bm{x}_i$ and $\bm{x}_j$, $\theta_i$ and $\theta_j$ is the angle between $\bm{f}_i$ \& $\bm{W}_{y_i}$ and $\bm{f}_j$ \& $\bm{W}_{y_j}$ , respectively.}
	\label{fig:quality_metric}
\end{figure}

Figure \ref{fig:quality_metric} shows the geometry interpretation of the 2D feature and two samples $(\bm{x}_i, y_i)$ and $(\bm{x}_j, y_j)$ with different difficulties to recognize. $\theta_j$ is bigger than $\theta_i$, which means that $\bm{x}_j$ is harder to recognize compared to $\bm{x}_i$. Consequently, the quality score of $\bm{x}_j$ is smaller than $\bm{x}_i$.

\subsection{Tiny Face Quality Network}\label{sec:tinyFQnet}

The primary purpose of learning-based FIQA is to provide a prediction function for the face image quality score. Unlike previous learning-based methods that access quality scores via deep or hand-crafted features, we adopt an end-to-end method to directly predict the quality score of a raw image through deep CNN. The reason is that a deep CNN can learn more relative features from raw images compared with features extracted from pre-trained models. 

We noticed a few works that adopt an end-to-end deep CNN \cite{zhuang2019recognition,hernandez2019faceqnet} to predict face image quality. 
% {\color{red} 
The core difference is that the proposed method in this work doesn't need to pick an image as a reference using other methods before generating quality scores, while \cite{hernandez2019faceqnet} needs to use ICAO compliance software to pick an image with the highest score and \cite{zhuang2019recognition} need to select an image as a  reference through human perceptual vision. Given this point, both \cite{hernandez2019faceqnet} and \cite{zhuang2019recognition} are more like reference-based methods, while our method is more similar to non-reference methods.%}
Besides, these works didn't take the memory and computation costs into account and used very complex networks (ResNet-50 in \cite{zhuang2019recognition} and VGG-16 in \cite{hernandez2019faceqnet}) as face quality networks. However, both computation and memory costs are key factors when one needs to apply FIQA to real FR scenarios.

\begin{figure}[!htb]
	\centering 
	\includegraphics[width=0.95\linewidth]{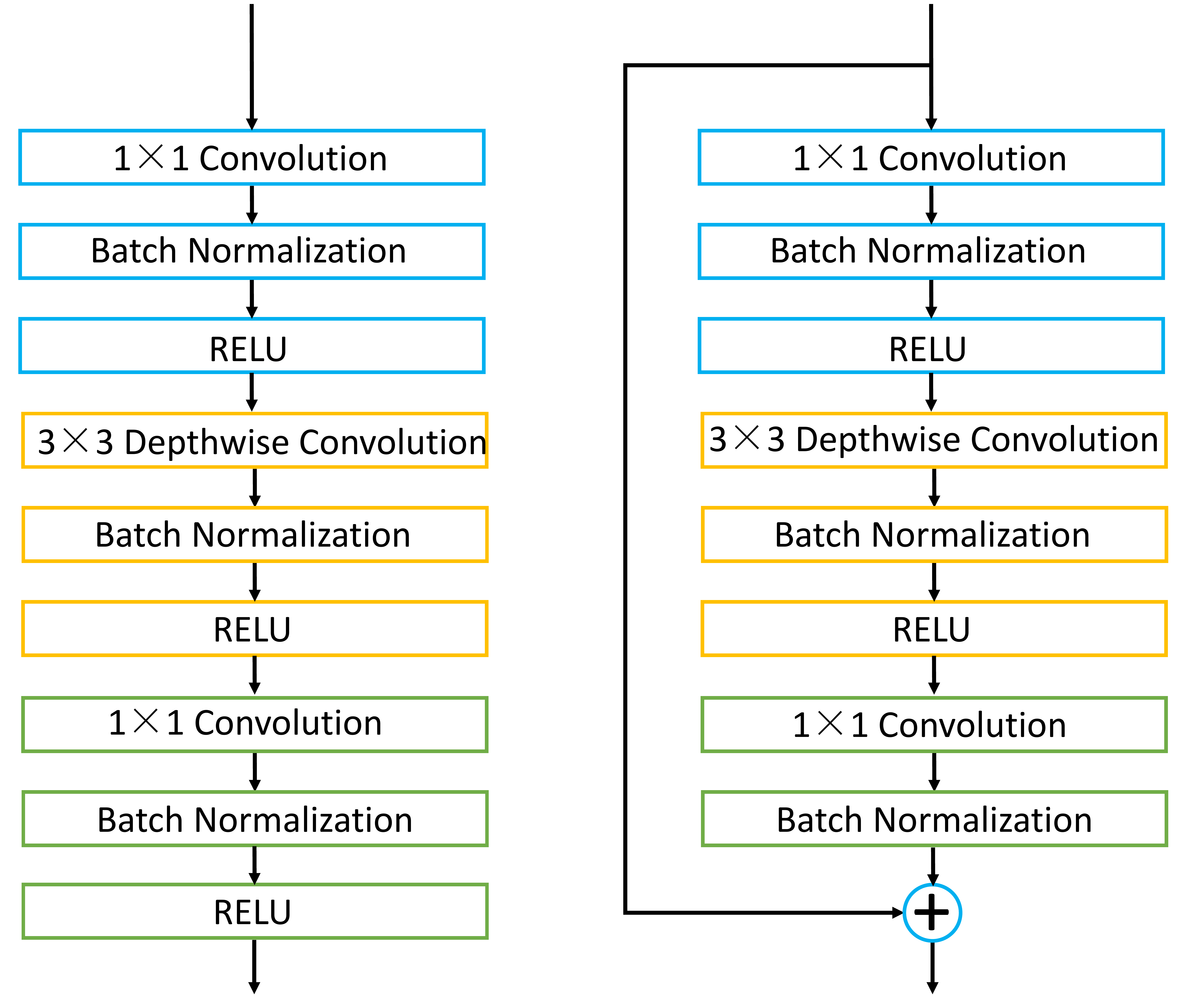}
	\put(-192,-8){(a)}
	\put(-55 ,-8){(b)}
	\caption{The details of two basic blocks in tinyFQNet. (a): non-residual block; (b): residual block.}
	\label{fig:block}
\end{figure}

In this work, we design a Tiny Face Quality Assessment Network (noted as tinyFQnet) that is tiny but efficient for FIQA. The design of the tinyFQnet is highly modularized by following MobileNetV2 \cite{Howard2017MobileNets}. It stacks two kinds of blocks shown in Figure \ref{fig:block}, both of which share the same topology but are only different in whether there is a residual connection. Each block has three convolutional layers, and each convolutional layer is followed by a batch normalization layer and a ReLU layer. Both blocks share the same hyper-parameters (such as filter sizes, padding, bias, etc.) except for the input channels and output channels.  

\begin{table}[htbp]
	\caption{ Our tinyFQnet architecture. Each conv layer is followed by a batch norm layer and a ReLU layer. Block1 and Block2 are shown in Fig \ref{fig:block}.}
	\label{FIQA_arch}
	\centering
        \setlength{\tabcolsep}{2mm}{
	\begin{tabular}{c c c}
		\hline
	layer name & output size & tinyFQnet                                                        \\
	\hline
	input      & 64 $\times$ 64 $\times$ 3     &                                                                     \\
	\hline
	conv    & 32 $\times$ 32 $\times$ 11    & 3 $\times$ 3 $\times$ 11,stride=2                                                     \\
	\hline
	Block1    & 32x32x2     & [\begin{tabular}[c]{@{}l@{}}1 $\times$ 1, 8\\ 3 $\times$ 3, 8\\ 1 $\times$ 1, 2\end{tabular}] $\times$ 1, stride=1    \\
	\hline
	Block2 & 16 $\times$ 16 $\times$ 5     & [\begin{tabular}[c]{@{}l@{}}1 $\times$ 1, 8\\ 3 $\times$ 3, 8\\ 1 $\times$ 1, 5\end{tabular}] $\times$ 1, stride=2    \\
	\hline
	Block1 & 16 $\times$ 16 $\times$ 5     & [\begin{tabular}[c]{@{}l@{}}1 $\times$ 1, 20\\ 3 $\times$ 3, 20\\ 1 $\times$ 1, 5\end{tabular}] $\times$ 1, stride=1  \\
	\hline
	Block2 & 8 $\times$ 8 $\times$ 11      & [\begin{tabular}[c]{@{}l@{}}1 $\times$ 1, 20\\ 3 $\times$ 3, 20\\ 1 $\times$ 1, 11\end{tabular}] $\times$ 1, stride=2 \\
	\hline
	Block1 & 8 $\times$ 8 $\times$ 11      & [\begin{tabular}[c]{@{}l@{}}1 $\times$ 1, 44\\ 3 $\times$ 3, 44\\ 1 $\times$ 1, 11\end{tabular}] $\times$ 2, stride=1 \\
	\hline
	Block1    & 8 $\times$ 8 $\times$ 22      & [\begin{tabular}[c]{@{}l@{}}1 $\times$ 1 $\times$ 44\\ 3 $\times$ 3 $\times$ 44\\ 1x1x22\end{tabular}]x1, stride=1    \\
	\hline
	conv    & 8 $\times$ 8 $\times$ 256     & 1 $\times$ 1 $\times$ 256, stride=1                                                            \\
	\hline
	avgpool    & 256         &                                                                     \\
	\hline
	fc         & 1           & 1 $\times$ 256                                                               \\
	\hline
	\end{tabular}}
\end{table}

Table \ref{FIQA_arch} shows the details of the architecture of tinyFQnet, noted as $Q$. There are $7$ blocks in $Q$, including $5$ non-residual blocks and $2$ residual blocks in $Q$.  The output of $Q$ will be normalized to (0,1) by a sigmoid layer. $Q$ only has  2.356 Mflops and 21.806k parameters, while ResNet-50 has 3.898 Gflops and 22.421M parameters, and VGG-16 has 14.528 Gflops and 128.041M parameters. Compared with ResNet-50 and VGG-16 used in previous methods, our tinyFQnet has a significant advantage in memory and computation costs. Thanks to the low computation cost, our tinyFQnet can significantly speed up FR systems.
To cooperate with the FR network, we use RetinaFace \cite{deng2019retinaface} to detect and align images. Then, the aligned images are resized to $64 \times 64$ which is the same as tinyFQnet's input size. L2 regression loss in Equation  \ref{equa:L2_loss} is adopted to train the tinyFQnet as follows:
\begin{equation}
	L=\frac{1}{N}\sum_{i=1}^{N}\left \| \phi(x_i,\theta) - q_i \right \|_2 , \label{equa:L2_loss}
	\end{equation}
% {\color{red} 
where $\phi(x,\theta)$ denotes the prediction model with parameter $\theta$.%} 
Stochastic gradient descent is used to train the network. The learning rate is initialized to 0.01 and degrades by 0.1 at every 5 epochs. The training batch size is set as 1024, and the weight decay is set as 0.0001. The training takes 17 epochs in total.
	
\subsection{
	Generating training dataset with quality label%} 
	}
	\label{sec:labelling_data_generation}

Usually, learning-based methods require amounts of labeled data to learn the desired function. 
Previous learning-based FIQA methods tried to train a prediction model from small lab-collected datasets (e.g., SCFace, FRGC, GBU, and Multi-PIE) due to the lack of a large number of face images that are labeled with quality scores. The lack of training data limits the capability of deep learning models. 
Similar to \cite{zhuang2019recognition} and \cite{hernandez2019faceqnet}, we adopt a feasible but fast method to generate a large number of labeled data. More specifically, since the proposed quality metric only involves a well-trained FR model and a training dataset, we can automatically generate amounts of data with quality labels and use labeled data to train our tinyFQnet. 

 The whole process of generating the labeled dataset for training the tinyFQnet is shown as follows:
\begin{algorithm}
    \caption{Quality label generating process}
    \begin{algorithmic}
    \STATE Train model $Q$ on $\mathcal{D}$;
    \STATE Extract weights $\mathcal{W}$ of the classifier layer in $Q$;
    \STATE Extract features $\mathcal{F}$ of training data $\mathcal{D}$ using $Q$;
    \STATE Compute quality scores of $\mathcal{D}$ using Equation \ref{equa:fiqa_metric};
    \end{algorithmic}
\end{algorithm}

\noindent The first step is to train the FR model $Q$. Usually, we can use an existing well-trained FR model and its training dataset. Then, we extract the weights $\mathcal{W}$ of the classifier and all training data features from the FR model. Finally, the quality of the image can be computed with Equation \ref{equa:fiqa_metric}.

\subsection{Data sampling and augmentation strategy for balancing the distribution of scores}\label{sec:smooth_strategy}

\begin{figure}[!htb]
	\centering 
	\includegraphics[width=0.95\linewidth]{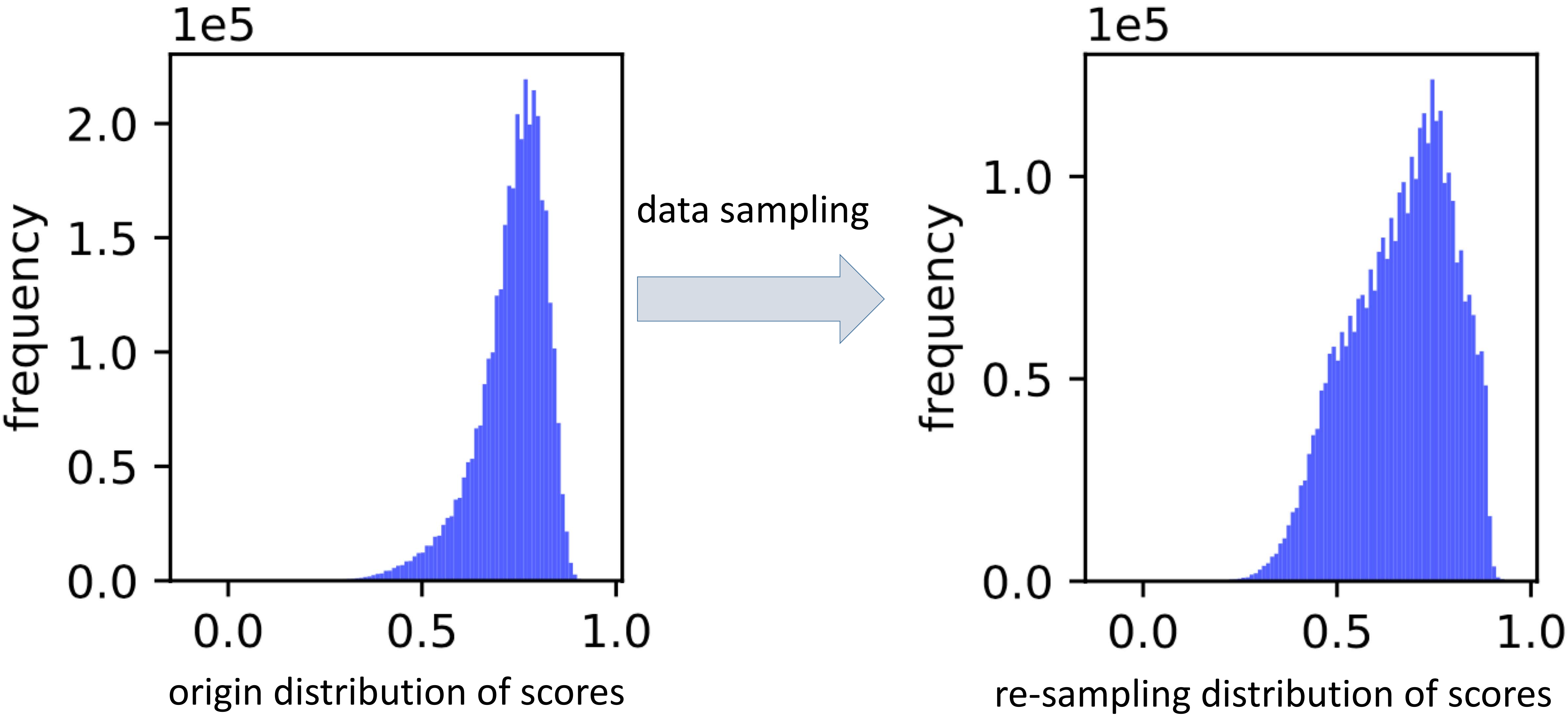}
	\caption{Distribution of quality scores under different data sizes, including large size(left column), middle size(middle column), and small size(right column). The dataset is Ms-Celeb-1M \cite{guo2016ms}, and the FR model is R50. The top row shows the results of a random sampling strategy, and the bottom row shows the results of a smooth sampling strategy.}
	\label{fig:score_balancing}
\end{figure}

Although we can generate amounts of data with the quality scores through the method above, we found that the quality score distribution is highly unbalanced. The left subfigure in Figure \ref{fig:score_balancing} shows the histogram distribution of quality scores generated by Section~\ref{sec:labelling_data_generation}. It can be observed that most images are concentrated in a narrow range of scores. The fact, that the proposed quality metric is directly related to the cosine similarity between the image and its class centers in embedding space, means that the lower the training loss of the FR model is, the closer the features of the intra-class are. Consequently, a well-performed FR model will lead to an unbalanced distribution of quality scores, which will mislead the optimized results of the tinyFQnet.

We adopt a data sampling strategy that samples the data by identity and scores. To be more specific, we first filter those identities of which the number of related images is less than a threshold (in our experiments, we set the threshold to be 100 for Ms-Celeb-1M dataset). By doing so, we get nearly 7,900 identities and 860 thousand images.
Then we divide images into 100 bins in a range from 0 to 1.0 according to quality scores with equal intervals. We over-sample those images of which quality scores range in the lowest 10\% and the highest 5\%, and downsample the remaining 80\% of images at each bin respectively (the sampling rate is dependent on the number of images in a bin). 
After applying identities filtering and scores sampling strategy, the distribution of quality scores in the training dataset is more balanced than before, as shown in the right subfigure of Figure \ref{fig:score_balancing}.

\section{ Experimental Results}\label{experiment}
  
\subsection{Experimental Setup}

  To evaluate the effectiveness of our FIQA on real FR systems, we consider the tiny recognition-oriented quality network (tinyRQNet) as a component independent of FR. We only use tinyRQNet to select the appropriate frames from a video. More specifically, we use the tinyRQNet to predict all the images in a video or a template (a template consists of several images and videos belonging to the same subject in the IJB-B dataset) and select the image with the highest quality score. Then, we evaluate those selected images using the FR model.

  We compare the proposed method with 
  % {\color{red}
  three learning-based methods, including SVM \cite{best2018learning}, RQS \cite{chen2015face} and FaceQnet \cite{hernandez2019faceqnet}.
  RQS learns to predict and rank the quality scores from hand-crafted features. SVM \cite{best2018learning} uses a support vector machine to predict the quality scores with features extracted by a deep FR model. Similar to our method, FaceQnet adopts deep CNN to predict quality scores from raw images.%}
  Beyond these comparisons, we also present the results of the most common perceptual factors, including blurriness, JPEG compression, head pose, and their combination. All the quality scores predicted by each method are rescaled to a range from 0 to 100, of which a higher score means higher quality.
    
  We use two classical networks, including a complicated network ResNet-50 \cite{he2016deep} and a lightweight network EfficientNet-b0 \cite{tan2019efficientnet}, to train the FR model on MS-Celeb-1M \cite{guo2016ms} separately. 
  % {\color{red} 
  ResNet-50 is widely used in many tasks as a baseline model due to its powerful capacity, while EfficientNet-b0 achieves considerably better accuracy compared to classical MobileNet-V2 \cite{Howard2017MobileNets} while using fewer parameters and flops.%}
  All the images are aligned using 5 landmarks provided by RetinaFace \cite{deng2019retinaface} and cropped to $112  \times 112$ resolution. For all learning-based methods, we keep the same settings as the original papers. All the training is carried out on 8 TiTANX GPUs.  Note that we focus on face image quality assessment in this work, rather than learning video-level representations for face recognition. One can apply the proposed method to video-level face recognition, but this is beyond the scope of our work.
   
\subsection{Datasets and Protocols}
% {\color{red}
Three popular video-based (or template-based) datasets are used for evaluation: IJB-B \cite{whitelam2017iarpa}, IJB-C \cite{maze2018iarpa}, and YTF \cite{Wolf2011Face}.%}
 The IJB-B dataset consists of 21,798 images and 7,011 videos from 1,845 subjects captured under unconstrained conditions. Instead of image-to-image or video-to-video recognition and verification, the IJB-B challenge protocol aims to evaluate the FR model on templates. Due to large variations existing in different templates, this protocol is more difficult for the FR model than other benchmarks. We follow the template-based 1:1 verification task to evaluate the proposed method against other methods on the IJB-B dataset. 
%  {\color{red}
 The IJB-C dataset adds an extra 1,686 subjects based on IJB-B and contains 31,334 images and 11,779 videos. To improve the representation of the global population, the IJB-C emphasizes occlusion and diversity of subjects. By increasing the size and variability of the dataset, the IJB-C is more challenging than the IJB-B and other datasets in unconstrained face recognition.%}
    
The YTF dataset is a video-based dataset with 3,425 videos belonging to 1,595 different subjects. Each subject has 2.15 videos on average. All videos are collected in unconstrained conditions with large variations in poses, expressions, illuminations, etc. Like the IJB-A verification task, the YTF protocol splits the dataset into ten-fold cross-validation sets. Each set contains 500 randomly selected video pairs, of which 250 pairs are positive and the other 250 pairs are negative.

\subsection{Visualization of different FIQA methods}
\begin{figure*}[htbp]
\centering 
\includegraphics[width=0.98\textwidth]{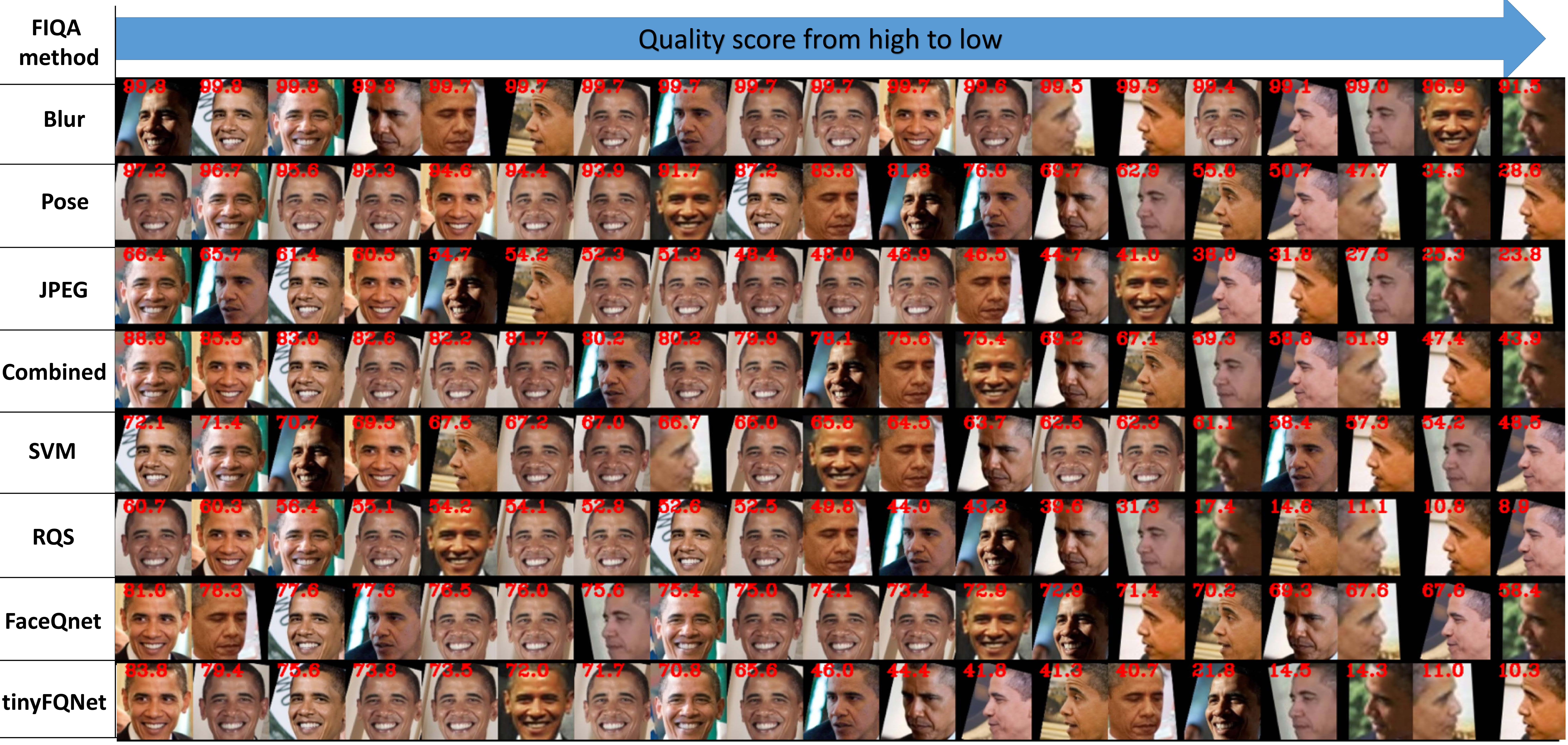}
\caption{Some face images randomly selected from one subject in the IJB-B dataset. We rank them with quality scores predicted by several FIQA methods.}\label{fig:scores_visualize}
\end{figure*}

Figure \ref{fig:scores_visualize} shows the images and corresponding quality scores predicted by different FIQA methods in the IJB-B dataset. We rank the face images by the quality scores from high to low. The first four methods are perceptual-based, and the last three methods are learning-based. The Blur only considers the blurriness of face images, and it chooses those with high resolution while ignoring other factors, such as head pose. The Pose only considers the head pose of a face. Although it can choose images under neural head poses, these images may be too blurry to recognize. The JPEG only considers the jpeg compression rate of face images, and the less compressed images are with higher quality scores. The combination consists of Blur, Pose, and JPEG. The learning-based methods learn quality prediction functions from labeled datasets. It shows the image with a higher quality score contains relatively more information about the identity and is easier to recognize in most situations.

\subsection{Memory and Computation Costs} \label{sec:memory_computation}

The memory cost and computation cost are the key factors when applying FIQA to a real FR system. Fewer parameters and less computation cost can improve the throughput of a real FR system. 

\begin{table}[htbp]
  \caption{The memory and computation costs of different methods. Since several methods are based on aligned images while the others are not, we also present whether it is necessary to align images.}\label{tab:memory_computation}%
  \centering
  \setlength{\tabcolsep}{1mm}{
    \begin{tabular}{c|c|c|c}
      \toprule
    method & need aligning? & \pbox{20cm}{parameters \\ (MB)} & \pbox{20cm}{computation\\(Gflops)} \\
    \midrule
    Blur  & no    & few   & low \\
    Pose  & yes   & 0     & low \\
    JPEG & no    & 0     & low \\
    % \hline
    SVM   & yes   & few   & low \\
    RQS   & yes   & few   & low \\
    FaceQNet & yes   & 23    & 3.9 \\
    tinyFQNet & yes   & 0.022 & 0.0024 \\
    \bottomrule
    \end{tabular}}
\end{table}%

% {\color{red} 
Table \ref{tab:memory_computation} shows the memory and computation costs of different methods. Generally speaking, perceptual-based methods need fewer parameters, while learning-based methods need more memory and computation costs. Although SVM has fewer parameters and low computation cost since it is based on the feature extracted from the FR model, it needs to process all images using the FR model regardless of that the image may be dropped, which would lower throughput. The RQS is based on handle-crafted features, including HoG, Gabor, Gist, LBP, and the feature of a face alignment network. Although less memory and low computation cost, the RQS is not easy to deploy on GPU devices. Compared with FaceQnet which needs nearly 23M parameters and 3.9 Gflops, tinyFQnet only has 21.8k parameters and about 0.0024 Gflops. Due to the efficient design of tinyFQNet, the proposed method is easy to serve as a plug-in with the FR model while the extra computation and memory cost are needed.   
% }

\subsection{Quantitative evaluation on %{\color{red}
IJB-B and IJB-C Datasets%}
} \label{sec:ijbb_result}
Table \ref{tab:IJBBresTab} summarizes the overall results of the 1:1 verification task on the IJB-B dataset. We use true positive rates (TPR) under $10^i$(i= -1, -2, -3, -4, -5) and false positive rates (FPR) as the evaluation metric. We choose the random selection as the baseline. We evaluate on two different recognition models, including ResNet-50 \cite{he2016deep} and EfficientNet-b0 \cite{tan2019efficientnet}.  We compare the tinyFQnet with three learning-based methods, including SVM \cite{best2018learning}, RQS \cite{chen2015face}, FaceQnet \cite{hernandez2019faceqnet}. All these methods showed higher performance compared with perceptual-based methods. Since both the \cite{chen2015face} and \cite{hernandez2019faceqnet} have released their code, we use the released code to compute the quality scores. For SVM \cite{best2018learning}, we implement their method and replace the training dataset with MS1M used in this paper. We also report the results of perceptual-based methods, including Blur, Pose, JPEG, and their combination.

\begin{table*}[]
  \caption{The overall results on the 1:1 verification task of the IJB-B dataset. Two different recognition models ResNet-50 \cite{he2016deep} and EfficientNet-b0 \cite{tan2019efficientnet} are chosen as recognition models. The true accept rates(TAR) vs. false positive rates(FAR) are used as the evaluation metric.} \label{tab:IJBBresTab}%
  \centering
  \setlength{\tabcolsep}{5mm}{
  \begin{tabular}{c|c|ccccc}
      \toprule
  \multirow{2}{*}{FR model}            & \multirow{2}{*}{method} & \multicolumn{5}{c}{tpr}                         \\
                                   &                         & fpr=e-1 & fpr=e-2 & fpr=e-3 & fpr=e-4 & fpr=e-5 \\
                                   \midrule
  \multirow{9}{*}{ResNet-50}       & random                  & 92.09   & 87.64   & 83.08   & 77.22   & 64.02   \\
                                   & Blur                    & 93.62   & 89.96   & 86.71   & 82.05   & 65.25   \\
                                   & Pose                    & 93.45   & 90.34   & 86.91   & 82.39   & 75.57   \\
                                   & JPEG                    & 95.27   & 93.14   & 90.87   & 88.46   & 81.93   \\
                                   & Combination             & 95.55   & 93.75   & 91.37   & 88.56   & 83.23   \\
                                   & SVM                     & 94.75   & 91.44   & 88.51   & 84.13   & 75.29   \\
                                   & RQS                     & 96.06   & 93.78   & 91.82   & 89.42   & 83.13   \\
                                   & FaceQnet                & 95.15   & 92.76   & 90.36   & 87.29   & 79.63   \\
                                  %  & aimall                  & 95.76   & 93.7    & 91.73   & 88.85   & 81.01   \\
 & \textbf{tinyFQNet}  & \textbf{95.99}   & \textbf{94.32}   & \textbf{92.73}   & \textbf{90.78}   & \textbf{85.21}   \\
                                   \midrule
 \multirow{9}{*}{EfficientNet-b0}  & random                  & 92.65   & 87.81   & 82.72   & 75.34   & 53.74   \\
                                   & Blur                    & 93.98   & 90.04   & 86.39   & 79.71   & 48.04   \\
                                   & Pose                    & 93.99   & 90.31   & 86.64   & 81.65   & 70.29   \\
                                   & JPEG                    & 95.7    & 93.15   & 90.99   & 87.85   & 79.06   \\
                                   & Combination             & 96.11   & 93.73   & 91.55   & 88.42   & 80.84   \\
                                   & SVM                     & 95.11   & 91.56   & 87.61   & 82.71   & 61.54   \\
                                   & RQS                     & 96.14   & 94.12   & 91.85   & 88.89   & 81.8    \\
                                   & FaceQnet                & 95.47   & 92.74   & 90.22   & 86.61   & 76.36   \\
                                  %  & aimall                  & 95.94   & 93.84   & 91.72   & 88.44   & 80.89   \\
 & \textbf{tinyFQNet} & \textbf{96.31}   & \textbf{94.41}   & \textbf{92.65}   & \textbf{90.16}   & \textbf{84.06}   \\
                                   \bottomrule
  \end{tabular}}
  \end{table*}

Table \ref{tab:IJBBresTab} shows that the learning-based methods outperform the perceptual-based methods, and both of them are superior to random selection. Among all these methods, our tinyFQnet achieves 85.21 tpr@fpr=e-5 when using the ResNet-50 network as the FR model (84.06 tpr@fpr=e-5 for EfficientNet-b0), which is the best performance. The SVM performs the worst among learning-based methods. A possible explanation for the results of SVM is that the goal of the FR model is to learn feature representation invariant to perceptual or other potential factors, and the feature of an image extracted from the FR model contains little information on image quality. SVM, which tries to learn a quality prediction function from the feature of a face image,  may not learn an effective prediction function about the image quality.

\begin{table*}[]
  \caption{The overall results on the 1:1 verification task of the IJB-C dataset. The true accept rates(TAR) vs. false positive rates(FAR) are used as the evaluation metric.}\label{tab:fiqa_ijbcres}%
  \centering
    \setlength{\tabcolsep}{5mm}{
  \begin{tabular}{c|c|ccccc}
    \toprule
  \multirow{2}{*}{FR model}            & \multirow{2}{*}{method} & \multicolumn{5}{c}{tpr}                         \\
                                   &                         & fpr=e-1 & fpr=e-2 & fpr=e-3 & fpr=e-4 & fpr=e-5 \\
                                   \midrule
  \multirow{9}{*}{ResNet-50}       & random                  & 92.24   & 87.88   & 83.11   & 76.75   & 68.42   \\
                                   & Blur                    & 94.33   & 90.69   & 87.32   & 83.28   & 74.65   \\
                                   & Pose                    & 93.8    & 90.95   & 87.45   & 83.06   & 75.22   \\
                                   & JPEG                    & 96      & 94.04   & 91.88   & 89.27   & 84.64   \\
                                   & Combination             & 95.97   & 93.79   & 91.41   & 88.81   & 83.77   \\
                                   & SVM                     & 94.94   & 91.98   & 88.92   & 84.67   & 77.21   \\
                                   & RQS                     & 96.17   & 94.21   & 92.13   & 89.72   & 85.36   \\
                                   & FaceQnet                & 95.37   & 93.12   & 90.8    & 87.34   & 82.3    \\
 & \textbf{tinyFQNet} & \textbf{96.29}   & \textbf{94.77}   & \textbf{93.05}   & \textbf{91.12}   & \textbf{87.68}   \\
                                   \midrule
  \multirow{9}{*}{EfficientNet-b0} & random                  & 92.31   & 87.66   & 82.97   & 75.51   & 51.6    \\
                                   & Blur                    & 94.25   & 90.66   & 87.41   & 81.3    & 57.88   \\
                                   & Pose                    & 94.45   & 91.09   & 87.16   & 82.13   & 69.76   \\
                                   & JPEG                    & 96.17   & 93.73   & 91.84   & 88.67   & 81.02   \\
                                   & Combination             & 96.21   & 93.61   & 91.31   & 88.23   & 80.23   \\
                                   & SVM                     & 95.29   & 91.92   & 88.1    & 82.97   & 68.78   \\
                                   & RQS                     & 96.46   & 94.35   & 92.21   & 89.37   & 84.54   \\
                                   & FaceQnet                & 95.49   & 93.13   & 90.61   & 86.8    & 80.12   \\
 & \textbf{tinyFQNet} & \textbf{96.53}   & \textbf{94.83}   & \textbf{93.05}   & \textbf{90.73}   & \textbf{86.06}   \\
                                   \bottomrule
  \end{tabular}}
  \end{table*}

% {\color{red} 
We also conduct experiments on a more recent and challenging benchmark IJB-C dataset \cite{maze2018iarpa}. Table \ref{tab:fiqa_ijbcres} summarizes the overall results of the 1:1 verification task on the IJB-C dataset. Similar to the results on IJB-B,  the learning-based methods outperform the perceptual-based methods, and the random selection strategy performs the worst.
% }

\begin{table*}[htbp]
  \caption{Performance evaluation for two recognition models with different FIQA methods on the YTF dataset. We adopt 10-folder cross-validation to calculate the verification accuracy. }\label{tab:YTFresTab}%
  \centering
  \begin{tabular}{cc|c|c}
    \toprule
  \multicolumn{2}{c}{method}       & accuracy with ResNet-50 & accuracy with Efficientnet-b0 \\
  \midrule
                   & random    & 96.34         & 96.53               \\
  perceptual-based & blur      & 96.39         & 96.53               \\
                   & pose      & 96.58         & 96.55               \\
                   & JPEG      & 96.72         & 96.72               \\
                   & combined  & 96.11         & 96.54               \\
                   \hline
  learning-based   & SVM       & 95.57         & 96.01               \\
                   & RQS       & 96.89         & \textbf{97.02}      \\
                   & FaceQnet  & 96.35         & 96.64               \\
                   & tinyFQNet & \textbf{97.02}& 97                  \\
                   \bottomrule     
  \end{tabular}
  \end{table*}

\subsection{Quantitative evaluation on YTF Dataset}
Table \ref{tab:YTFresTab} shows the results on the YTF dataset. Similar to the results shown in Table \ref{tab:IJBBresTab}, our tinyFQnet achieves the best performance among all methods. We find that learning-based methods slightly outperform perceptual-based methods. Unlike the results on the IJB-B dataset, the random selection achieves a comparable performance with the others on the YTF dataset. This is probably due to the fewer variations in a YTF video compared to an IJB-B template. Among all the methods, SVM performs the worst, as explained in Section \ref{sec:ijbb_result}. An interesting phenomenon in Table \ref{tab:YTFresTab} is that the combination of Blur, Pose, and JPEG performs even worse than each separate method.

\subsection{Ablation Studies}

\subsubsection{The impact of FIQA network's size}

Since the memory cost and computation cost are the key factors when applying FIQA to a real FR system, we explore the impact of the size of a network on performance in this section. 
More specifically, we use two different networks: MobileNetV2 and ResNet-50. There are four different architectures for MobileNetV2, including mbn\_t4\_w0.35\_64, mbn\_t6\_w1\_64, mbn\_t4\_w0.35\_112, and mbn\_t6\_w1\_112, where $t$ represents the channels expansion and w represents the width multiplier for basic block in MobileNetV2, the postfix number represents the input size of network.

\begin{table*}[htbp]
  \caption{Results of different networks on the IJB-B dataset. Five architectures, including mbn\_t4\_w0.35\_64, mbn\_t6\_w1\_64, mbn\_t4\_w0.35\_112, mbn\_t6\_w1\_112, and ResNet-50 are explored.} \label{tab:network_influence}
  \centering
  \setlength{\tabcolsep}{7mm}{
  \begin{tabular}{c|c|c|c|c}
  \toprule
  network        & input size & parameters & (Mflops)  & tpr@fpr=e-5 \\ 
  \midrule
  mobilenet\_t4\_w0.35 & 64x64       & 0.013M     & 1.43  & 85.21       \\ 
  mobilenet\_t6\_w1    & 64x64       & 0.02M      & 2.5   & 85.81       \\ 
  mobilenet\_t4\_w0.35 & 112x112     & 0.5M       & 44.4  & 85.19       \\ 
  mobilenet\_t6\_w1    & 112x112     & 2.1M       & 206.9 & 86.38       \\
  ResNet-50      & 112x112     & 22.4M      & 3907  & 86.93       \\ 
  \bottomrule
  \end{tabular}}
  \end{table*}

Table \ref{tab:network_influence} shows the results of different networks on IJB-B. As illustrated in the table, for a particular FR model, the size of FIQA network makes little difference on FR's performance. On the IJB-B dataset, the largest network ResNet-50 with 23M parameters achieves 86.93 tpr@fpr=e-5, only 1.7 higher than the smallest network mbn\_t4\_w0.35\_64 with 21.8k parameters. On the YTF dataset, the size of FIQA network almost makes no difference on the FR's performance.

\subsubsection{The impact of sampling strategy under different data sizes} \label{sec:sampling_strategy}
\begin{figure}[!htb]
	\centering 
	\includegraphics[width=0.95\linewidth]{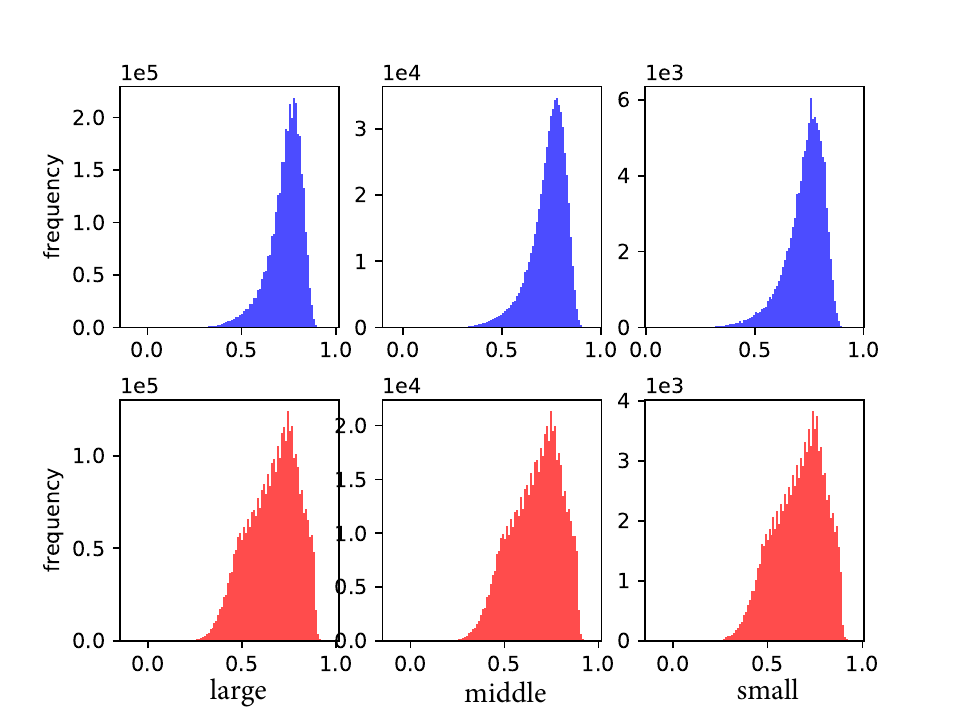}
	\caption{The distribution of quality scores under different data sizes, including large, middle, and small. The top row shows the results of the random sampling strategy, and the bottom row shows the results of a smooth sampling strategy.}
	\label{fig:score_distribution}
\end{figure}

To evaluate the impact of the data sampling strategy, we evaluate the tinyFQNet on FR performance when using different data sampling strategies to generate a training dataset. Two different sampling strategies are chosen for comparison, including the random strategy and the smooth strategy, as mentioned in ~\ref{sec:smooth_strategy}.
Figure \ref{fig:score_distribution} shows the histogram distribution of quality scores when applying two different strategies, the top row for the random strategy and the bottom row for the smooth strategy.

Table \ref{tab:distribution_results} shows the results of tinyFQNet on the IJB-B data set when using different sampling strategies to generate a training dataset for training tinyFQNet. For a fair comparison, we keep the same number of training images for both two sampling strategies. It can be observed that the number of identities of the smooth strategy is much lower than the random strategy due to applying identity filtering. In all three different data sizes, the smooth strategy is better than the random strategy.

  \begin{table*}[htbp]
    \caption{ Results on IJB-B dataset when using two different sampling strategies for generating training datasets to train tinyFQNet. To make the results more convincing, we conduct this experiment on three datasets with different sizes.}
  \label{tab:distribution_results}
  \centering
  \setlength{\tabcolsep}{7mm}{
    \begin{tabular}{c|c|c|c}
      \toprule
    sampling & data size & number(images, ids) & IJBB tpr@fpr=e-5 \\
    \midrule
    random          & large   & 3.5M, 92.9K       & 83.95            \\
    random          & middle  & 0.61M, 87.4K       & 84.73            \\
    ranom           & small   & 0.11M, 57.0K       & 84.78            \\
    % \midrule
    smooth          & large   & 3.5M, 34K         & 85.85            \\
    smooth          & middle  & 0.61M, 5.3K        & 85.64            \\
    smooth          & small   & 0.11M, 0.9K        & 85.45            \\
    \bottomrule
    \end{tabular}}
    \end{table*}

\subsubsection{The influence of the labeling model}
Since the FR model generates quality scores, a natural thought is how the labeling model (we denote the model used to generate quality scores as the labeling model.) would influence the overall performance of an FR model.

\begin{table*}[htbp]
  \centering
  \caption{Results of different quality labelling models on the IJB-B dataset.}\label{tab:label_model}
  \setlength{\tabcolsep}{7mm}{
  \begin{tabular}{c|c|c|c}
  \toprule
  FR model        & labeling model  & IJBB tpr@fpr=e-5 & YTF 10-fold accuracy \\ 
  \midrule
  ResNet-50       & ResNet-50       & 85.21            & 97.0           \\ 
  ResNet-50       & EfficientNet-b0 & 85.15            & 97.0                 \\ 
  EfficientNet-b0 & EfficientNet-b0 & 83.71            & 96.9                 \\ 
  EfficientNet-b0 & ResNet-50       & 84.06            & 97.0                 \\ 
  \bottomrule
  \end{tabular}}
  \end{table*}

Table \ref{tab:label_model} shows the results of two different labeling models (used to generate quality labels for face images to train tinyFQnet) on the performance of two different FR models. It shows that both FR models perform better when using ResNet-50 as a labeling model to generate quality scores, but there is no significant difference between them. It can be concluded that with a better labeling model, the FIQA model can improve the performance of an FR system.

\section{Conclusion}\label{conclusion}
   
In this paper, we present a novel deep FIQA method, in which a novel and effective recognition-oriented metric is proposed that directly links face image quality assessment with the performance of the face recognition model. We design a tiny but efficient face quality network (tinyFQnet) and use it to train a FIQA model. We provide analysis by many experiments about the influences of network size on FR's performance and show that even a tiny network with only 21.8K parameters can achieve comparable performance to a much more complicated network with 23M parameters. We show that learning-based methods are superior to perceptual-based methods, and the proposed method outperforms other learning-based methods on two classical benchmarks.
However, limitations exist. The quality metric relies only on cosine similarity comparisons and could explore alternatives and needs to test on more diverse datasets to confirm generalizability. Further efforts on more diverse datasets, alternate metrics, and model optimizations can build on this initial exploration of linking quality directly to recognition accuracy.

%-------------------------------------------------------------------------

{\small
\bibliographystyle{cvm}
\bibliography{cvmbib}
}

\end{document}